\documentclass[10pt,twocolumn,letterpaper]{article}

\usepackage{cvpr}              

\usepackage[dvipsnames]{xcolor}

\definecolor{cvprblue}{rgb}{0.21,0.49,0.74}
\usepackage[pagebackref,breaklinks,colorlinks,citecolor=cvprblue]{hyperref}

\usepackage{multirow}
\usepackage{makecell}
\usepackage[ruled]{algorithm2e}
\usepackage{mathtools}
\usepackage{amssymb}
\usepackage{amsmath}
\usepackage[accsupp]{axessibility}

\def\paperID{7869} 
\def\confName{CVPR}
\def\confYear{2024}

\title{Structure Matters: Tackling the Semantic Discrepancy \\ in Diffusion Models for Image Inpainting}
\author{Haipeng Liu$^{1}$\quad Yang Wang$^{1}$\thanks{Yang Wang is the corresponding author.}\quad Biao Qian$^{1}$\quad Meng Wang$^{1}$\quad Yong Rui$^{2}$\\
$^{1}$Hefei University of Technology, China\quad $^{2}$Lenovo Research, China\\
{\tt\small hpliu\_hfut@hotmail.com, yangwang@hfut.edu.cn,}\\
{\tt\small \{hfutqian,eric.mengwang\}@gmail.com, yongrui@lenovo.com}
}

\begin{document}

\makeatletter
\newcommand{\rmnum}[1]{\romannumeral #1}
\newcommand{\Rmnum}[1]{\expandafter\@slowromancap\romannumeral #1@}
\makeatother

\maketitle

\begin{abstract}
Denoising diffusion probabilistic models (DDPMs) for image inpainting aim to add the noise to the texture of the image during the forward process and recover the masked regions with the unmasked ones of the texture via the reverse denoising process. Despite the meaningful semantics generation, the existing arts suffer from the semantic discrepancy between the masked and unmasked regions, since the semantically dense unmasked texture fails to be completely degraded while the masked regions turn to the pure noise in diffusion process, leading to the large discrepancy between them.
In this paper, we aim to answer how the unmasked semantics guide the texture denoising process; together with how to tackle the semantic discrepancy, to facilitate the consistent and meaningful semantics generation. To this end, we propose a novel structure-guided diffusion model for image inpainting named \textbf{StrDiffusion}, to reformulate the conventional texture denoising process under the structure guidance to derive a simplified denoising objective for image inpainting, while revealing:  1) the semantically sparse structure is beneficial to tackle the semantic discrepancy in the early stage, while the dense texture generates  the reasonable semantics in the late stage;  2) the semantics from the unmasked regions essentially offer the time-dependent structure guidance for the texture denoising process, benefiting from the time-dependent sparsity of the structure semantics.
For the denoising process, a structure-guided neural network is trained to estimate the simplified denoising objective by exploiting the consistency of the denoised structure between masked and unmasked regions. Besides, we devise an adaptive resampling strategy as a formal criterion as whether the structure is competent to guide the texture denoising process, while regulate their semantic correlations. Extensive experiments validate the merits of StrDiffusion over the state-of-the-arts. Our code is available at \textup{\url{https://github.com/htyjers/StrDiffusion}}.
\end{abstract}


\begin{figure}[t]
  \centering
  \setlength{\belowcaptionskip}{-1.5cm}
    \setlength{\abovecaptionskip}{0.2cm}
  \includegraphics[width=0.9\linewidth]{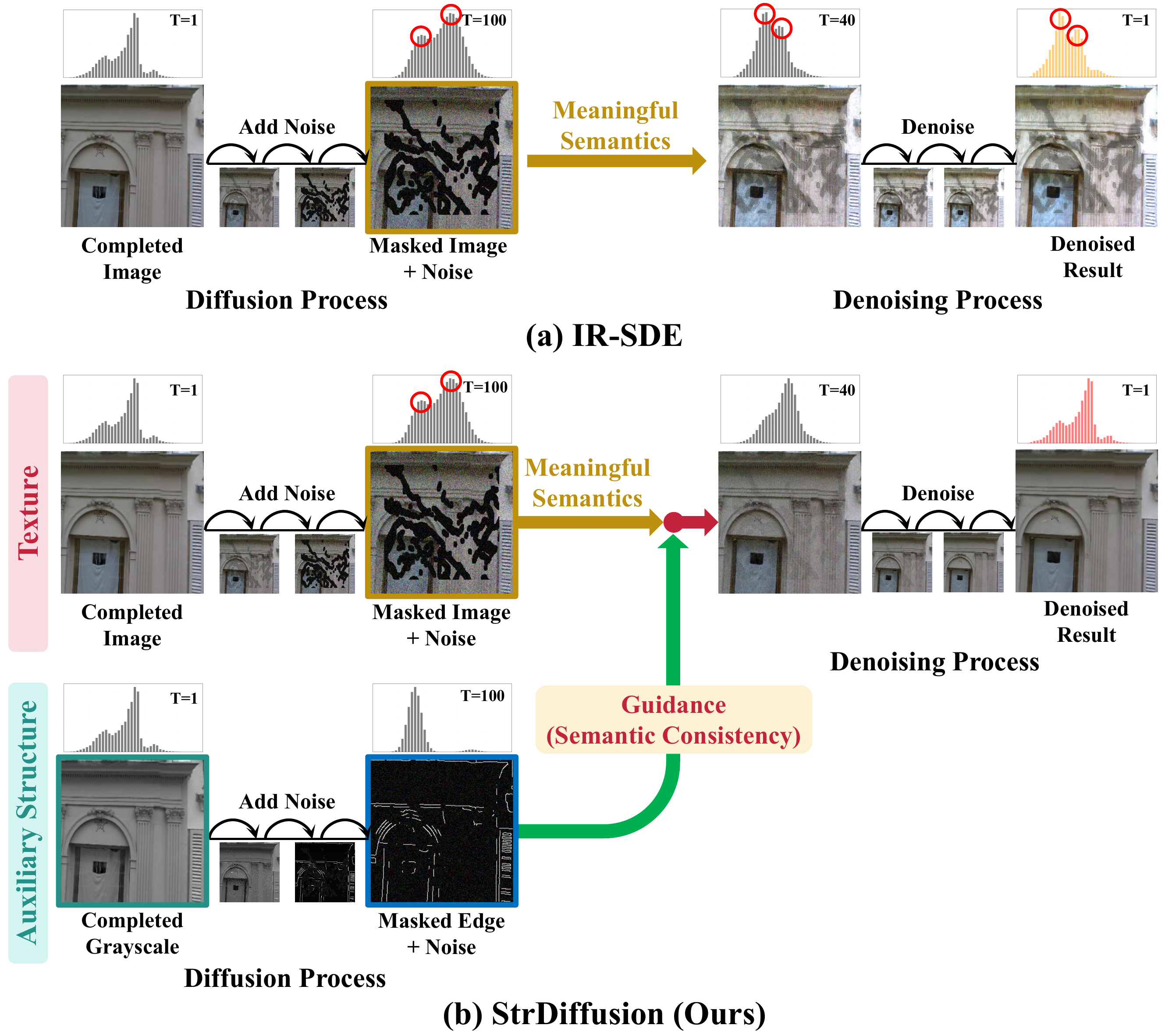}
  \caption{Existing arts, \emph{e.g.}, IR-SDE \cite{luo2023image} (a), suffer from the semantic \emph{discrepancy} (\textcolor[RGB]{255,0,0}{$\bigcirc$}) between the masked and unmasked regions despite of the meaningful semantics for the masked regions during the denoising process. Our StrDiffusion (b) tackles the semantic discrepancy issue via the \emph{guidance} of the auxiliary \emph{sparse} structure, yielding the consistent and meaningful denoised results. The experiments are conducted on PSV \cite{doersch2012makes}.}
  \label{comparison_sota}
\end{figure}

\begin{figure*}[t]
  \centering
  \setlength{\belowcaptionskip}{-0.5cm}
    \setlength{\abovecaptionskip}{0.2cm}
  \includegraphics[width=\textwidth]{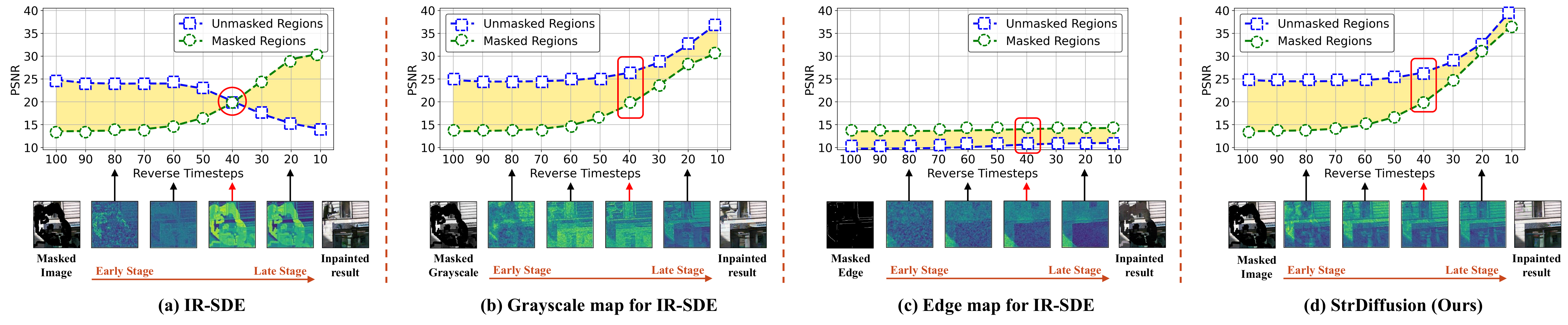}
  \caption{Illustration of the motivating experiments about whether the sparse structure is beneficial to alleviating the discrepancy issue during the denoising process for image inpainting. Apart from the dense texture for IR-SDE \cite{luo2023image} (a), the unmasked semantics combined with the Gaussian noise is further set as the sparse structure, \emph{e.g.}, the grayscale map (b) and edge map (c). Our StrDiffusion (d) can tackle the semantic discrepancy via the progressively sparse structure. The \textcolor[RGB]{189,145,14}{shadow area} indicates the discrepancy between the masked and unmasked regions during the denoising process. The PSNR (higher is better) reflects the recovered semantics for the masked (unmasked) regions compared to the completed image (\emph{i.e.}, ground truth) by calculating the semantic similarity between them. The inpainted results are obtained by combining the masked regions of denoised results with the original masked images.}
  \label{motivating_exper}
\end{figure*}

\section{Introduction}
\label{sec:intro}
Recently, image inpainting has supported a wide range of applications, \emph{e.g.}, photo restoration and image editing, which aims to recover the masked regions of an image with the semantic information of unmasked regions, where the \emph{principle} mainly covers two aspects: the \emph{reasonable semantics} for the masked regions and their \emph{semantic consistency} with the unmasked regions.
The prior work mainly involves the diffusion-based \cite{bertalmio2000image, efros2001image} and patch-based \cite{barnes2009patchmatch, darabi2012image, hays2007scene} scheme, which tends to inpaint the small mask or repeated patterns of the image via simple color information, while fails to handle irregular or complicated masks.
To address such issue, substantial attention \cite{xie2019image,zeng2021generative, li2020recurrent,li2022misf} has shifted to the convolutional neural networks (CNNs), which devote themselves to encoding the local semantics around the masked regions, while ignore the global information from the unmasked regions, incurring that the regions far away from the mask boundary remain burry. Recently, self-attention mechanism \cite{yan2019PENnet, yi2020contextual, li2022mat, deng2022hourglass, Ko_2023_ICCV} is proposed to globally associate the masked regions with unmasked ones in the form of the segmented image patches, enhancing the semantic consistency between them. However, such strategies suffer from poor semantic correlation between the varied patches inside the masked regions. To this end, the semantically sparse structure is exploited \cite{liu2022delving, yu2022unbiased, Liao_2021_CVPR, Liu2019MEDFE, wang2021parallel, Guo_2021_ICCV, dong2022incremental, cao2022learning, yu2021diverse, wan2021high} to strength their correlations, which, however, implies the heavy reliance on the semantic consistency between the structure and texture, hence inevitably bears the artifact in the inpainted output.

\begin{figure*}[t]
  \centering
  \setlength{\belowcaptionskip}{-0.5cm}
    \setlength{\abovecaptionskip}{0.2cm}
  \includegraphics[width=0.85\textwidth]{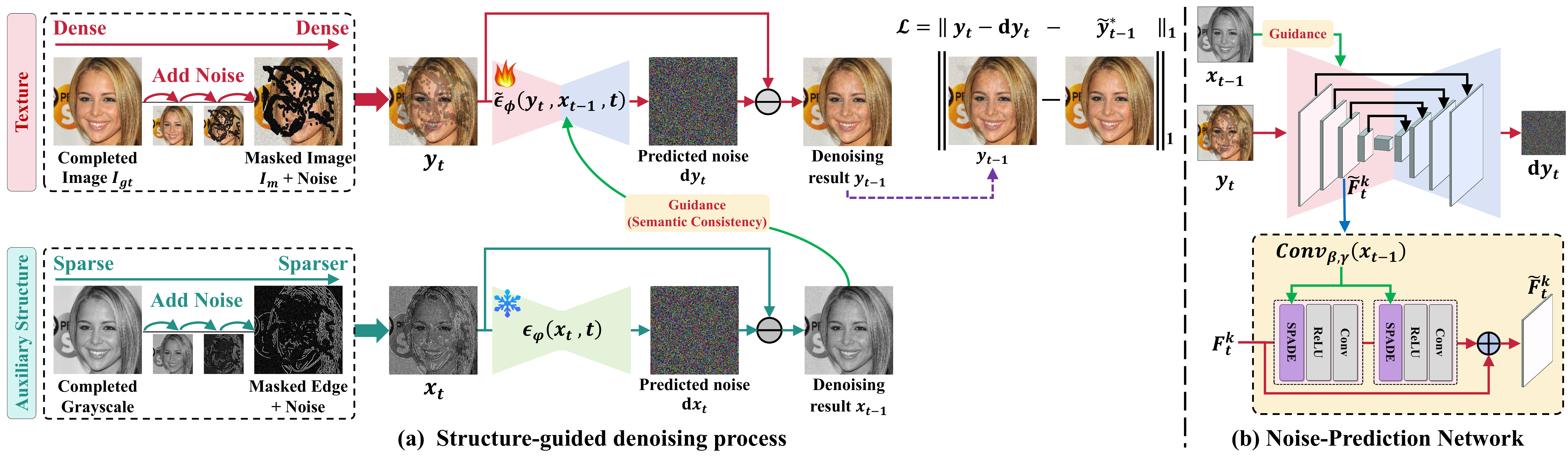}
  \caption{Illustration of the proposed StrDiffusion pipeline. Our basic idea is to tackle the semantic discrepancy between masked and unmasked regions via the guidance of the progressively sparse structure (a), which guides the texture denoising network (b) to generate the consistent and meaningful denoised  results.  }
  \label{architecture}
\end{figure*}

Fortunately, the denoising diffusion probabilistic models (DDPMs) \cite{ho2020denoising, song2020score} have emerged as the powerful generative models to deliver remarkable gains in the semantics generation and mode converge, hence well remedy the poor semantics generation for image inpainting \cite{wang2022zero, zhu2023designing, lugmayr2022repaint,luo2023image}. Specifically, instead of focusing on the training process, \cite{lugmayr2022repaint} proposed to adopt the pre-trained DDPMs as the prior and develop a resampling strategy to condition the reverse denoising process during the inference process. Further, \cite{luo2023image} attempted to model the  diffusion process for image inpainting by exploiting the semantics from the unmasked regions, yielding an optimal reverse solution to the denoising process. These methods mostly exhibit the semantically meaningful inpainted results with the advantages of DDPMs, whereas \emph{overlook} the semantic consistency between the masked and unmasked regions.  The intuition is that the semantically dense \textit{unmasked} texture is degraded into the combination of the unmasked texture and the Gaussian noise, while the \textit{masked} regions turn to the pure noise during the diffusion process, leading to the large discrepancy between \textit{them} (see Fig.\ref{comparison_sota}(a) and Fig.\ref{motivating_exper}(a)), and \emph{motivates} the followings:

\begin{itemize}
    \item[(1)] \emph{How does the unmasked semantics guide the texture denoising process for image inpainting?} Our motivating experiments (see Fig.\ref{motivating_exper}) suggest that, when the unmasked semantics combined with the noise turns to be sparser, \emph{e.g.}, utilizing the grayscale or edge map of the masked image as an alternative, the discrepancy issue is largely alleviated, meanwhile, bears the large semantic information loss in the inpainted results; see Fig.\ref{motivating_exper}(b)(c). Therefore, the invariable semantics from the unmasked regions over time fail to guide the texture denoising process well.
    \item[(2)] Following (1), one naturally wonders \emph{how to yield the denoised results with consistent and meaningful semantics.} It is apparent that the sparse structure benefits the semantic consistency in the early stage while the dense texture tends to generate the meaningful semantics in the late stage during denoising process, implying a balance between the consistent and reasonable semantics of the denoised results. To further yield the desirable results, we take into account the guidance of the structure as an auxiliary over the texture denoising process; see Fig.\ref{comparison_sota}(b).
\end{itemize}

To answer the above questions, we propose a novel structure-guided texture diffusion model, named StrDiffusion, for image inpainting, to tackle the semantic discrepancy between masked and unmasked regions, together with the reasonable semantics for masked regions; technically,  we reformulate the conventional texture denoising process under the guidance of the structure to derive a \emph{simplified} denoising objective for image inpainting. For the denoising process, a structure-guided neural network is trained to estimate the simplified denoising objective during the denoising process, which exploits the time-dependent consistency of the denoised structure between the masked and unmasked regions to mitigate the discrepancy issue. Our StrDiffusion \emph{reveals}: 1) the semantically \emph{sparse} structure encourages the consistent semantics for the denoised results in the early stage, while the \emph{dense} texture carries out the semantic generation in the late stage; 2) the semantics from the unmasked regions essentially offer the time-dependent structure guidance to the texture denoising process, benefiting from the time-dependent sparsity of the structure semantics. Concurrently, we remark that whether the structure guides the texture well greatly depends on the semantic correlation between them. As inspired, an adaptive resampling strategy comes up to monitor the semantic correlation and regulate it via the resampling iteration.
Extensive experiments on typical datasets validate the merits of StrDiffusion over the state-of-the-arts.

\section{Structure-Guided Texture Diffusion Models}
\label{sec:formatting}
Central to our method lies in three aspects: 1) how to tackle the semantic discrepancy between the masked and unmasked regions (Sec.\ref{tdim}); 2) structure-guided denoising network for image inpainting (Sec.\ref{estimate_denoising}); and 3) more insights on whether the structure guides the texture well (Sec.\ref{further_insight}).
Before shedding light on our technique, we elaborate the DDPMs for image inpainting.


\subsection{Preliminaries: Denoising Diffusion Probabilistic Models for Image Inpainting} \label{prelimin}
Given a completed image $I_{gt} \in \mathbb{R}^{3 \times H \times W}$ and the binary mask $M \in \{0,1\}^{1\times H \times W}$ (0 for masked regions, 1 for unmasked regions), image inpainting aims to recover the masked image $I_{m}=I_{gt} \odot M$ towards a inpainted image. For diffusion models, the inpainted results are obtained by merging the masked regions of denoised results with the original masked images. The typical DDPMs for inpainting \cite{luo2023image} basically involve the forward texture diffusion process and reverse texture denoising process:

\noindent \textbf{Forward Texture Diffusion Process.}
During the forward diffusion process, the semantic information from unmasked regions is utilized to guide the texture denoising process for image inpainting. Specifically, for the diffusion process with $T$ timesteps,  we set the initial texture state $y_{0} = I_{gt}$ and the terminal texture state $y_{T}$ as the combination of  the masked image $\mu_{y}=I_{m}$ and the Gaussian noise $\varepsilon$. For any state $t \in [0,T]$, the diffusion process $\{y_t\}_{t=0}^T$ via the mean-reverting stochastic differential equation (SDE) \cite{song2021maximum} is defined as
\begin{small}
\begin{equation}
\label{eq:sde_f_tex}
\begin{split}
  \mathrm{d}y = \theta_{t}(\mu_{y}-y)dt+\eta_{t}\mathrm{d}w,
  \end{split}
\end{equation}
\end{small}
where $\theta_{t}$ and $\eta_{t}$ are time-dependent positive parameters that characterize the speed of the mean-reversion and the stochastic volatility during the diffusion process, which are constrained by $\eta_{t}^2/\theta_{t}=2\lambda^2$ for the stationary variance since $\lambda$ is fixed as the noise level applied to $y_{T}$.  In particular, $\theta_{t}$ can be adjusted to construct different noise schedules for the texture; $w$ is a standard Wiener process \cite{song2020score} (known as Brownian motion), which brings the randomness to the differential equation.

\noindent \textbf{Reverse Texture Denoising Process.}
To recover the completed image $y_{0}$ from the terminal texture state $y_{T}$, we reverse the diffusion process by simulating the SDE backward in time, formulated as
\begin{small}
\begin{equation}
\label{eq:reverse_tex}
\begin{split}
  \mathrm{d}y =[\theta_{t}(\mu_{y}-y)-\eta_{t}^{2}\nabla_{y}\log q_{t}(y)]\mathrm{d}t+\eta_{t}\mathrm{d}\hat{w},
  \end{split}
\end{equation}
\end{small}
where $\hat{w}$ is a reverse-time Wiener process and $q_{t}(y)$ stands for the marginal probability density function of texture $y_t$ at time $t$. The score function $\nabla_{y}\log p_{t}(y)$ is unknown, which can be approximated by training a conditional time-dependent neural network ${\epsilon}_{\phi}(y_{t},t)$ \cite{ho2020denoising}.
Alternatively, we find the optimal reverse texture state $y_{t-1}^{*}$ from $y_{t}$ in ($t-1$)-th timestep via the maximum likelihood learning, which is achieved by minimizing the negative log-likelihood:
\begin{small}
\begin{equation}
\label{reverse_texture}
\begin{split}
    y_{t-1}^{*}&=\arg\min_{y_{t-1}} [-\log q(y_{t-1}|y_{t},y_{0})].\\
    \end{split}
\end{equation}
\end{small}
By solving the above objective, we have:
\begin{small}
\begin{equation}
\label{eq:texture1}
\begin{split}
    y_{t-1}^{*}=& \frac{1- e^{-2\overline{\theta}_{t-1}}}{1- e^{-2\overline{\theta}_{t}}}e^{-\theta_{t}^{'}}  (y_{t}-\mu_{y})\\
    &+\frac{1- e^{-2\theta_{t}^{'}}}{1- e^{-2\overline{\theta}_{t}}}e^{-\overline{\theta}_{t-1}}{(y_{0}-\mu_{y})} + {\mu_{y}},
    \end{split}
\end{equation}
\end{small}
where $\theta_{t}^{'}=\int_{t-1}^{t}\theta_{i}\mathrm{d}i$ and $\overline{\theta}_{t}=\int_{0}^{t}\theta_{z}\mathrm{d}z$. Thus, we can optimize ${\epsilon}_{\phi}$ via the following training objective:
\begin{small}
\begin{equation}
\label{}
\begin{split}
 \mathcal{L}^{t}_{\beta}(\phi)=\sum^{T}_{t=1}\beta_{t}\mathbb{E}\left[\|y_{t}-(\mathrm{d} y_{t})_{{\epsilon}_{\phi}}-y_{t-1}^{*}\|_p\right],
  \end{split}
\end{equation}
\end{small}
where $\beta_{t}$ is positive weight; $(\mathrm{d} y_{t})_{{\epsilon}_{\phi}}$ denotes the reverse-time SDE in Eq.(\ref{eq:reverse_tex}) and its score is predicted via the noise network ${\epsilon}_{\phi}(y_{t},t)$. $||\cdot||_p$ stands for the $\ell_p$ norm.

\begin{figure}
  \centering
    \setlength{\abovecaptionskip}{0.2cm}
  \setlength{\belowcaptionskip}{-0.5cm}
  \includegraphics[width=0.9\linewidth]{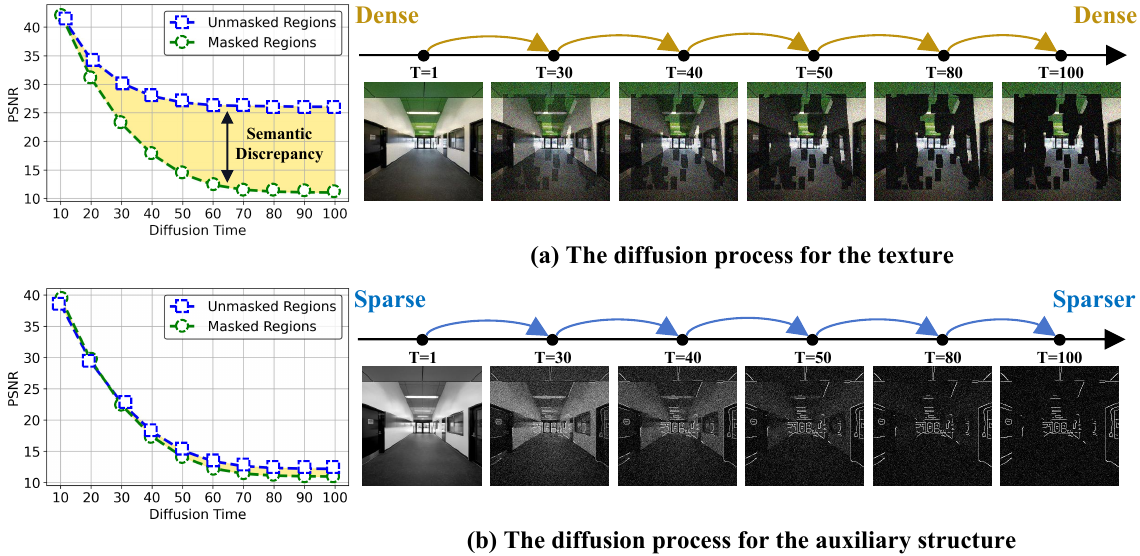}
  \caption{Illustration of the diffusion process for the dense texture (a) and sparse structure (b). In particular, the semantic sparsity of the structure is strengthened over time.}
  \label{progressive_sparsity}
\end{figure}

\subsection{Tackling the Semantic Discrepancy in Diffusion Models for Image Inpainting}\label{tdim}
Based on the above, the existing diffusion models for image inpainting \cite{luo2023image} suffer from the semantic discrepancy between the masked and unmasked regions, owing to the \emph{dense} semantics of the texture.
As opposed to the texture, the previous work \cite{liu2022delving,yu2022unbiased} suggests that the structure are generally considered to be \emph{sparse}, which motivates us to tackle the discrepancy issue via the structure.

\subsubsection{Structure Matters: Sparse Semantics benefits the Semantic Consistency}
\label{structure_matter}
To validate such intuition, we design a series of motivation experiments over \cite{luo2023image}, where we replace the masked image $\mu_{y}$ in the terminal texture state $y_T$   with the masked image $\mu_{x}$ from the \textit{sparse} structure, \emph{e.g.}, the grayscale map and edge map.
Our motivation experiments (see Fig.\ref{motivating_exper}) reveal the followings:
\begin{itemize}
    \item[(1)] With the dense semantics, the texture performs well in producing meaningful semantics, while suffers from the large semantic discrepancy between the masked and unmasked regions; see Fig.\ref{comparison_sota}(a) and Fig.\ref{motivating_exper}(a).
    \item[(2)] \emph{The semantically sparse structure benefits the consistent semantics}, while incurs the semantics loss in the inpainted results; see the grayscale and edge map version of \cite{luo2023image} in Fig.\ref{motivating_exper}(b)(c).
\end{itemize}
The above observations indicate that, unlike \cite{luo2023image} that fixes the texture guidance from the unmasked regions, the structure guidance from the unmasked regions \emph{varies} to the texture over time; meanwhile, \emph{merely} involving the semantics of either dense  texture or sparse structure fails to obtain the desirable inpainted results, since it actually attempts to perform the trade-off between the consistent and meaningful semantics. Motivated by that, we propose to exploit structure as an \emph{auxiliary}, which exhibits the sparser semantics than the texture, so as to tackle the discrepancy issue, as remarked below.

\noindent \textbf{Remark 1. Strengthening the semantic sparsity of the structure over time.}
Throughout the texture diffusion process, the semantic discrepancy between the masked and unmasked regions \emph{progressively} increases as the timestep goes towards to a totally masked status (see Fig.\ref{progressive_sparsity}(a)). To decrease the semantic discrepancy, we aim to strengthen the semantic sparsity of the structure over time, as opposed to the texture, during diffusion process.   Concretely, the \emph{sparse} grayscale map is degraded into the combination of the \emph{sparser} masked edge map and the Gaussian noise via the noise schedule; see Fig.\ref{progressive_sparsity}(b).

\emph{Our basic idea is to tackle the semantic discrepancy issue via the guidance of the structure during the denoising process, thus simplify the texture denoising objective, {i.e.}, the optimal solution (the ground truth) to the denoising process}, which is discussed in the next.

\subsubsection{Optimal Solution to the Denoising Process Under the Guidance of the Structure}
\label{optimal_solution}

As per the above observations, during the texture diffusion process, the distribution of the added noise in each timestep is disturbed by the semantic discrepancy between the masked and unmasked regions, incurring the difficulty to estimate such noise distribution via a noise-prediction network during the denoising process.
Instead of focusing on texture, we exploit both structure and texture for diffusion process, where the sparsity of structure alleviates the semantic discrepancy, and the texture attempts to preserve the meaningful semantic. Unlike the diffusion process to simultaneously exploit structure and texture, for  denoising process, the structure \emph{firstly} generates the consistent semantics ($x_{t-1}$) between masked and unmasked regions, which \emph{subsequently} guides the texture to generate the meaningful semantics ($y_{t-1}$); see Fig.\ref{architecture}. By the assistance of structure for StrDiffusion, the texture diffusion process generates both consistent and meaningful inpainting output.

\noindent \textbf{Remark 2.}  Different from the typical DDPMs for inpainting  \cite{luo2023image,lugmayr2022repaint}, our StrDiffusion exploits the semantic consistency of the denoised structure to guide the texture denoising process, yielding both consistent and meaningful denoised results. \emph{In particular}, we view the structure as an auxiliary to guide the texture rather than the inverse; the reason is two-fold: 1) during the inference denoising process, the masked regions mainly focus on the texture to be inpainted rather than the structure; 2) the structure focuses on the consistency between the masked and unmasked regions, lacking the meaningful semantics, hence still relies on the texture for inpainting.

Thanks to the guidance of the structure, the reverse transition for the texture in Eq.(\ref{reverse_texture}) can be reformulated via Bayes’ rule as follows:
\begin{small}
\begin{equation}
\label{denoising_1}
\begin{split}
    q(y_{t-1}|y_{t},x_{t-1})&=\frac{q(y_{t-1}|y_{t})q(x_{t-1}|y_{t-1},y_{t})}{q(x_{t-1}|y_{t})}\\
    &=\frac{q(y_{t}|y_{t-1})q(y_{t-1})q(x_{t-1}|y_{t-1})}{q(x_{t-1},y_{t})},
    \end{split}
\end{equation}
\end{small}
where the texture $y_{t}$ and $y_{t-1}$, together with the structure $x_{t-1}$, are actually conditioned on their initial states, \emph{i.e.}, $y_{0}$ and $x_{0}$, during the diffusion process, hence $q(y_{t-1})=q(y_{t-1}|y_{0})$ and $q(x_{t-1},y_{t})=q(x_{t-1})q(y_{t}|x_{t-1})=q(x_{t-1}|x_{0})q(y_{t}|x_{t-1},y_{0})$. Eq.(\ref{denoising_1}) can thus be simplified as
\begin{small}
\begin{equation}
\label{likelihood_obj}
\begin{split}
    &q(y_{t-1}|y_{t},y_{0},x_{t-1},x_{0})\\
    &= \frac{q(y_{t}|y_{t-1})q(x_{t-1}|y_{t-1})}{q(y_{t}|x_{t-1},y_{0})} \times \frac{q(y_{t-1}|y_{0})}{q(x_{t-1}|x_{0})}\\
    &\approx \frac{q(y_{t-1}|y_{0})}{q(x_{t-1}|x_{0})},\\
    \end{split}
\end{equation}
\end{small}
where $q(y_{t}|y_{t-1}) \approx q(y_{t}|x_{t-1},y_{0})$ and $q(x_{t-1}|y_{t-1})\approx 1$, since the texture $y_{t}$ can be obtained via the conditioned auxiliary structure $x_{t-1}$ during the diffusion process and initial texture $y_{0}$ as Remarked above. Based on Eq.(\ref{likelihood_obj}), the optimal reverse state is naturally acquired by minimizing the negative log-likelihood (\emph{see the supplementary material for more inference details}):
\begin{small}
\begin{equation}
\label{}
\begin{split}
    \tilde{y}_{t-1}^{*}&=\arg\min_{y_{t-1}} [-\log q(y_{t-1}|y_{t},y_{0},x_{t-1},x_{0})]\\
    &=\arg\min_{y_{t-1}} [-\log \frac{q(y_{t-1}|y_{0})}{q(x_{t-1}|x_{0})}],
    \end{split}
\end{equation}
\end{small}
where $\tilde{y}_{t-1}^{*}$ denotes the ideal state reversed from $\tilde{y}_{t}$ under the structure guidance.
To solve the above objective, we compute its gradient and setting it to be zero:
\begin{small}
\begin{equation}
\label{eq:ideal1}
\begin{split}
    &\nabla_{\tilde{y}_{t-1}^{*}} \{-\log q(\tilde{y}_{t-1}^{*}|y_{t},y_{0},x^{*}_{t-1},x_{0})\}= \\
     &-\nabla_{\tilde{y}_{t-1}^{*}}\log q(\tilde{y}_{t-1}^{*}|y_{0})+\nabla_{x_{t-1}^{*}}\log q(x_{t-1}^{*}|x_{0})=0,
    \end{split}
\end{equation}
\end{small}
where $x_{t-1}^{*}$ is the ideal state reversed from $x_{t}$ for the structure, similar in spirit to Eq.(\ref{eq:texture1}), given as
\begin{small}
\begin{equation}
\label{eq:ideal2}
\begin{split}
    x_{t-1}^{*}=& \frac{1- e^{-2\overline{\delta}_{t-1}}}{1- e^{-2\overline{\delta}_{t}}}e^{-\delta_{t}^{'}}{(x_{t}-\mu_{x})}\\
    &+\frac{1- e^{-2\delta_{t}^{'}}}{1- e^{-2\overline{\delta}_{t}}}e^{-\overline{\delta}_{t-1}}{(x_{0}-\mu_{x})}+ {\mu_{x}},
    \end{split}
\end{equation}
\end{small}
where structure $\mu_{x}$ is the masked version of its initial state $x_{0}$ and $\delta_{t}$ is time-dependent parameter that characterizes the speed of the mean-reversion. To simplify the notation, we have $\delta_{t}^{'}=\int_{t-1}^{t}\delta_{i}\mathrm{d}i$, $\overline{\delta}_{t-1}=\int_{0}^{t-1}\delta_{z}\mathrm{d}z$ and $\overline{\delta}_{t} = \overline{\delta}_{t-1} + \delta_{t}^{'}$. Since Eq.(\ref{eq:ideal1}) is linear, we can derive $\tilde{y}_{t-1}^{*}$ as
\begin{small}
\begin{equation}
\label{eq:texture2}
\begin{split}
    \tilde{y}_{t-1}^{*}=&\underbrace{  \left(\frac{1- e^{-2\overline{\theta}_{t-1}}}{1- e^{-2\overline{\delta}_{t}}}e^{-\delta_{t}^{'}}\right)   (x_{t}-\mu_{x})}_{\text{Consistency term for masked regions}}\\
    &-\underbrace{ \left(\frac{1- e^{-2\overline{\theta}_{t-1}}}{1- e^{-2\overline{\delta}_{t}}}e^{-\delta_{t}^{'}}\right)e^{-\overline{\delta}_{t}}  (x_{0}-\mu_{x})}_{\text{Balance term for masked regions}}\\
    & +\underbrace{e^{-\overline{\theta}_{t-1}} (y_{0}-\mu_{y})}_{\text{Semantics  term for masked regions}} + \underbrace{\mu_{y}}_{\text{Unmasked regions}}.
    \end{split}
\end{equation}
\end{small}

\noindent \textbf{Remark 3.} Eq.(\ref{eq:texture2}) implies that the ideal reverse state $\tilde{y}_{t-1}^{*}$ focuses more on the semantic consistency based on the semantic generation for image inpainting, to alleviate the semantic discrepancy issues. Under the guidance of the sparse structure, $\tilde{y}_{t-1}^{*}$ tends to maintain the consistency between masked and unmasked regions in the early stage of the denoising process, while recovering the reasonable semantics in the late stage.
Specifically, $\tilde{y}_{t-1}^{*}$ is mainly composed of four parts: 1) the consistency term  for the structure $x_{t}-\mu_{x}$ is utilized to form the masked regions inside the denoised texture and keep the semantic \emph{consistency} with the unmasked regions; 2) the semantics term for the texture $y_{0}-\mu_{y}$ progressively provides \emph{meaningful} semantics for masked regions as the denoising process progresses; 3) the negative balance term $x_{0}-\mu_{x}$ adaptively surpasses the semantic information from $y_{0}-\mu_{y}$ in the early stage and the consistency information from $x_{t}-\mu_{x}$ in the late stage; 4) the unmasked texture $\mu_{y}$ retains the semantics of the \emph{unmasked} regions for the denoised texture.
All four parts together constitute the \emph{optimal solution} to the structure-guided denoising process in the $(t-1)$-th timestep, \emph{i.e.}, the ground truth to optimize the noise-prediction network, thus yielding the denoised results with the consistent and desirable semantics (see Fig.\ref{inpainting_process}), as discussed in the next section.

\subsection{Structure-Guided Denoising Process: How does the Structure Guide the Texture Denoising Process?}
\label{estimate_denoising}
With the above optimal solution (\emph{i.e.}, the ground truth) in Eq.(\ref{eq:texture2}), the goal of the structure-guided denoising process is to estimate it via a noise-prediction neural network ${\tilde{\epsilon}}_{\phi}(y_{t},x_{t-1},t)$ under the guidance of the structure, where ${\tilde{\epsilon}}_{\phi}(y_{t},x_{t-1},t)$ is trained to predict the mean of the noise with the fixed variance of noise \cite{song2020score}.
Unlike the conventional texture denoising process in \cite{luo2023image},  the structure denoising network is pre-trained as an \emph{auxiliary} to obtain the denoised structure, \emph{e.g.}, $x_{t-1}$ in the $t$-th timestep.
To fulfill the structure guidance, a \emph{multi-scale spatially adaptive normalization strategy} is devised to condition a time-dependent noise network ${\tilde{\epsilon}}_{\phi}(y_{t},x_{t-1},t)$ for the texture, which deploys the spatially adaptive normalization and denormalization strategy \cite{park2019semantic} to incorporate the statistical information over the feature map across varied layers; see Fig.\ref{architecture}. Formally, given $x_{t-1}$ from a pre-trained structure denoising network, we achieve the structure guidance by reconstructing the feature map on the $k$-th layer of CNNs in time-dependent noise network $\tilde{\epsilon}_{\phi}$ for the texture (\emph{i.e.}, ${F}^{t}_{k} \in \mathbb{R}^{C_{k} \times H_{k} \times W_{k}}$) as
\begin{small}
\begin{equation}
\begin{aligned}
        & \tilde{F}^{t}_{k} = conv_{\gamma}\left(x_{t-1}\right)\frac{{F}^{t}_{k} - \mu_{k}^{t}}{\sigma_{k}^{t}}+conv_{\beta}\left(x_{t-1}\right),\\
        & \mu^{t}_{k}(h_k,w_k) = \frac{1}{C_{k}}\sum^{C_{k}}_{c_k=1}F^t_{k}(h_k,w_k, c_k),\\
        & \sigma^{t}_{k}(h_k,w_k) = \sqrt{\frac{1}{C_{k}}\sum^{C_{k}}_{c_k=1}(F^t_{k}(h_k,w_k, c_k)-\mu^{t}_{k}(h_k,w_k))^2}, \\
        &h_k=1,2,...,H_{k}, w_k=1,2,...,W_{k},
\end{aligned}
    \label{}
\end{equation}
\end{small}
where $conv_{\gamma}(\cdot)$ and $conv_{\beta}(\cdot)$ denote the mappings that convert the input $x_{t-1}$ to the scaling and bias values.  $\mu^{t}_{k}(h_k,w_k)$ and $\sigma^{t}_{k}(h_k,w_k)$ are the statistical mean and variance of the pixels across different channels at the position $(h_k,w_k)$. Based on that, we turn to optimize the noise network ${\tilde{\epsilon}}_{\phi}(y_{t},x_{t-1},t)$ to estimate the optimal solution $\tilde{y}_{t-1}^{*}$ (\emph{i.e.}, the ground truth) to the denoising process.
To this end, the overall training objective is formulated as
\begin{small}
\begin{equation}
\label{training_obj}
\begin{split}
 \mathcal{L}^{t}_{\tilde{\beta}}(\phi)=\sum^{T}_{t=1}\tilde{\beta}_{t}\mathbb{E}[\|\underbrace{y_{t}-(\mathrm{d} y_{t})_{{\tilde{\epsilon}}_{\phi}}}_{\text{Denoised} \ y_{t-1}} - \tilde{y}_{t-1}^{*}\|_p],
  \end{split}
\end{equation}
\end{small}
where $\tilde{\beta}_{t}$ is positive weight; $(\mathrm{d} y_{t})_{{\tilde{\epsilon}}_{\phi}}$ denotes the estimated reverse noise by the noise-prediction network. By minimizing Eq.(\ref{training_obj}), the inpainted results with consistent and meaningful semantics can be generated during the structure-guided denoising process.

\subsection{More Insights on Eq.(\ref{training_obj})}
\label{further_insight}
To validate the effectiveness of the guidance from the structure to the texture denoising process, we provide more insights on Eq.(\ref{training_obj}) by studying the semantic correlation between the denoised texture and structure during inference.

\subsubsection{Correlation Measure via a Discriminator}
\label{discriminator}
Apart from the semantic structure sparsity, the semantic correlation between the structure and texture is crucial to achieve the semantic consistency between the masked and unmasked regions.
To measure such semantic correlation, we adopt a discriminator network $D$ to yield the correlation score $D(y_{t-1},x_{t-1},t-1)$ (abbreviated as $D(y_{t-1},x_{t-1})$),  where the denoised texture $y_{t-1}$ and structure  $x_{t-1}$, together with the timestep $t-1$, serve  as the inputs.
To this end, we present a discriminator loss and a triplet loss to optimize $D$, as described below:

\noindent \textbf{Discriminator Loss.} During the denoising process, to provide the desirable guidance for the texture, the current structure $x_{t-1}$ is expected to be closely correlated with the texture $y_{t-1}$ in the current timestep, instead of the previous one $y_{t}$. Hence, we aim to discriminate $y_{t-1}$ and $y_{t}$ via a discriminator loss below:
\begin{small}
\begin{equation}
\label{loss1}
\begin{split}
    \mathcal{L}_{dis} =& -\mathbb{E}_{y_{t-1}}\left[\log\left(D(y_{t-1},x_{t-1})\right)\right] \\
    &-\mathbb{E}_{y_{t}}\left[\log\left(1-D(y_{t},x_{t-1})\right)\right].
  \end{split}
\end{equation}
\end{small}

\begin{figure}[t]
  \centering
  \setlength{\belowcaptionskip}{-0.5cm}
    \setlength{\abovecaptionskip}{0.2cm}
  \includegraphics[width=0.9\linewidth]{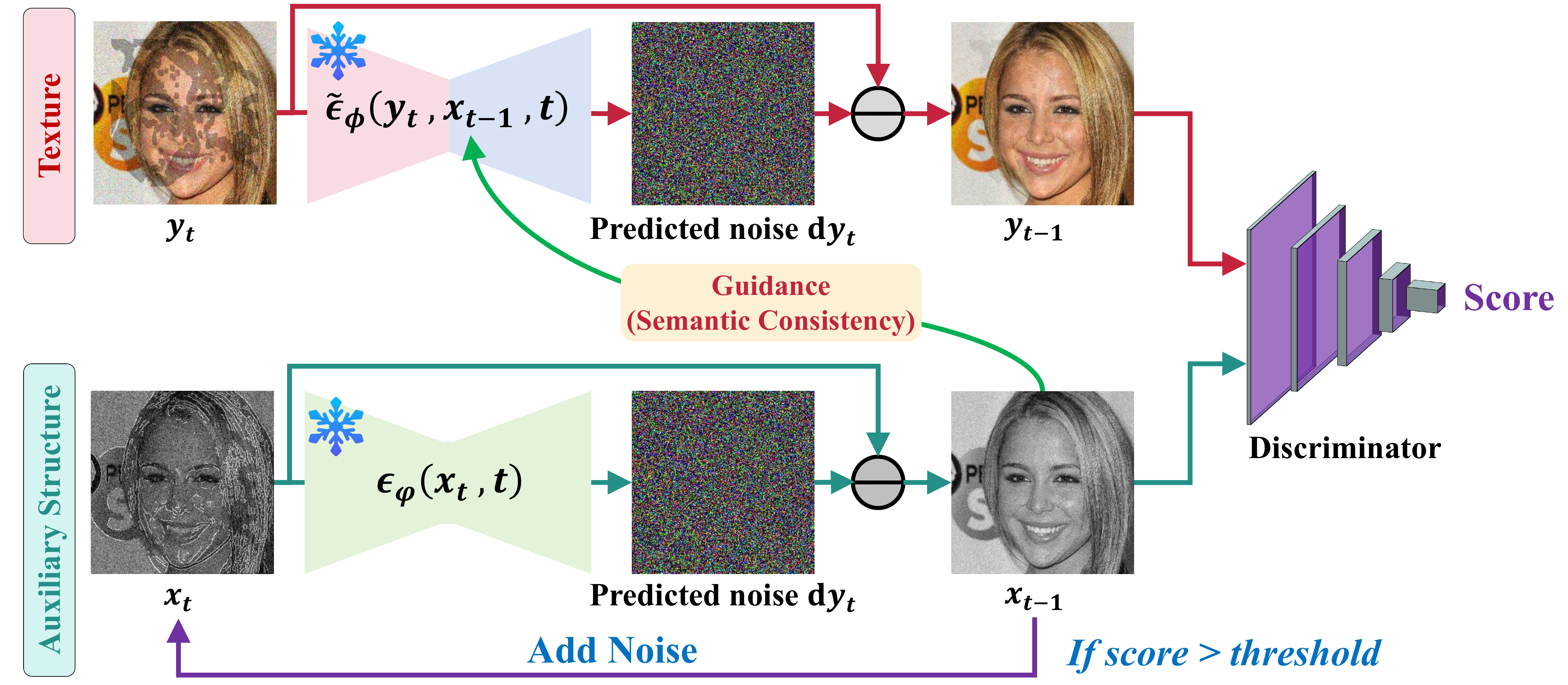}
  \caption{Illustration of the adaptive resampling strategy, which adaptively regulates the semantic correlation between the denoised texture and structure according to the score from the discriminator. }
  \label{adaptive_resampling}
\end{figure}

\begin{figure*}[t]
  \centering
    \setlength{\abovecaptionskip}{0.2cm}
\setlength{\belowcaptionskip}{-0.2cm}
  \includegraphics[width=0.9\linewidth]{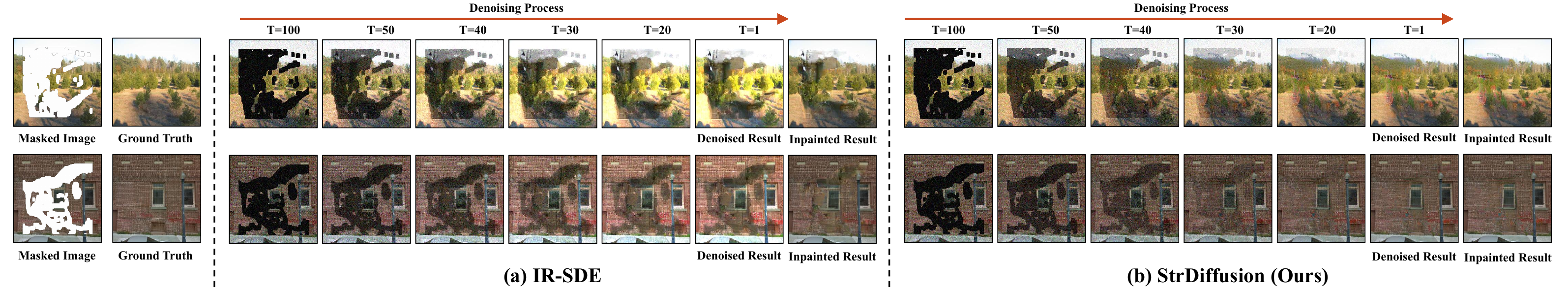}
  \caption{Illustration of the denoised results for IR-SDE \cite{luo2023image} (a) and StrDiffusion (b) in the varied timesteps during the denoising process.}
  \label{inpainting_process}
\end{figure*}

\begin{table*}[t]
    \centering
    \caption{Comparison of quantitative results (\emph{i.e.}, PSNR, SSIM \cite{wang2004image} and FID \cite{heusel2017gans}) under varied mask ratios on Places2 with irregular mask dataset. $\uparrow$: Higher is better; $\downarrow$: Lower is better; - : no results are reported. The percentage (\%) represents the ratios of varied irregular masked regions to the completed image. The best results are reported with \textbf{boldface}.}
    \label{tab:compare1}
    \resizebox{0.8\textwidth}{1.7cm}{
    \begin{tabular}{@{}lc||cccc|cccc|cccc@{}}
    \toprule
	\multicolumn{2}{c}{\textbf{Metrics}} & \multicolumn{4}{c}{\textbf{PSNR$\uparrow$}}  &\multicolumn{4}{c}{\textbf{SSIM$\uparrow$}}& \multicolumn{4}{c}{ \textbf{FID$\downarrow$} }\\ \cmidrule(r){1-2} \cmidrule(r){3-6} \cmidrule(r){7-10} \cmidrule(r){11-14}
     \textbf{Method} &\textbf{Venue} &\textbf{10-20\%} & \textbf{20-30\%} &\textbf{30-40\%} &\textbf{40-50\%} &\textbf{10-20\%} & \textbf{20-30\%} &\textbf{30-40\%} &\textbf{40-50\%} &\textbf{10-20\%} & \textbf{20-30\%} &\textbf{30-40\%} &\textbf{40-50\%} \\ \cmidrule(r){1-1} \cmidrule(r){2-2} \cmidrule(r){3-3} \cmidrule(r){4-4} \cmidrule(r){5-5} \cmidrule(r){6-6}   \cmidrule(r){7-7} \cmidrule(r){8-8} \cmidrule(r){9-9} \cmidrule(r){10-10}   \cmidrule(r){11-11} \cmidrule(r){12-12} \cmidrule(r){13-13} \cmidrule(r){14-14}
    \textit{RFR} \cite{li2020recurrent} &CVPR' 20 &27.26 &24.83 &22.75 &21.11 &0.929 &0.891 &0.830 &0.756 &17.88 &22.94 &30.68 &38.69 \\\hline
    \textit{PENNet} \cite{yan2019PENnet} &CVPR' 19 &26.78&23.53 &21.80&20.48  &0.946 &0.894 &0.841 &0.781 &20.40 &37.22 &53.74 &67.08\\
    \textit{HiFill} \cite{yi2020contextual} &CVPR' 20 &27.50 &24.64 &22.81 &21.23 &0.954 &0.914 &0.873 &0.816 &16.95 &29.30 &42.88 &57.12\\
    \textit{FAR} \cite{cao2022learning} &ECCV' 22 &28.36 &{25.48} &23.60 &22.18 &0.942 &0.903 &{0.861} &{0.818} &- &- &- &- \\
    \textit{HAN} \cite{deng2022hourglass} &ECCV' 22 &28.93 &25.44 &23.06 &21.38 &{0.957} &{0.903} &0.839 &0.762 &12.01 &{20.15} &{28.85} &{37.63}\\
    \textit{CMT} \cite{Ko_2023_ICCV} &ICCV' 23 &29.28&25.88&23.56&21.70&0.960&0.924&0.885&0.841&-&-&-&-\\\hline
    \textit{MEDFE} \cite{Liu2019MEDFE} &ECCV' 20 &28.13 &25.02 &22.98 &21.53 &0.958 &0.919 &0.874 &0.825 &14.41 &27.52 &38.45 &53.05\\
    \textit{CTSDG} \cite{Guo_2021_ICCV} &ICCV' 21  &28.91 &25.36 &22.94 &21.21 &0.952 &0.901 &0.834 &0.755 &15.72 &27.88 &42.44 &57.78 \\
    \textit{ZITS} \cite{dong2022incremental} &CVPR' 22 &28.31 &25.40 &23.51 &22.11 &0.942 &0.902 &0.860 &0.817 &- &- &- &- \\\hline
    \textit{RePaint*} \cite{lugmayr2022repaint} &CVPR' 22 &29.16 &26.02 &24.01 &22.11 &0.967 &0.933 &0.915 &0.862 &11.16 &20.62 &28.49 &41.16\\
    \textit{IR-SDE} \cite{luo2023image} &ICML' 23 &28.98 &25.37 &23.93  &21.68 &0.961 &0.928 &0.908  &0.864 &12.03 &21.34 &29.37 &39.77\\\hline \hline
    \textbf{StrDiffusion (Ours)} &-  &\textbf{29.45} &\textbf{26.28} &\textbf{24.43} &\textbf{22.37} &\textbf{0.973} &\textbf{0.941} &\textbf{0.917} &\textbf{0.874} &\textbf{8.750} &\textbf{16.16} &\textbf{23.72} &\textbf{36.93}\\
    \bottomrule
    \end{tabular}
  \label{table:compare1}
  }
  \vspace{-0.5cm}
\end{table*}

\noindent \textbf{Triplet Loss.}
Since the unmasked regions almost possess much denser semantics information compared to the masked ones during the denoising process, to well evaluate the correlation, Eq.(\ref{loss1}) is encouraged to focus more on the unmasked regions, which is achieved via a triplet loss:
\begin{small}
\begin{equation}
\label{loss2}
\begin{split}
    \mathcal{L}_{tri}=&\max(\|D(\bar{y}_{t-1},x_{t-1})-D(y_{t-1},x_{t-1}))\|_{2} \\
    &-\|D(\bar{y}_{t-1},x_{t-1})-D(y_{t},x_{t-1})\|_{2}+\alpha,0),
  \end{split}
\end{equation}
\end{small}
where $\bar{y}_{t-1} = y_{t-1}\odot M+y_{t}\odot \left(1-M\right)$ is constructed to associate the unmasked regions of $y_{t-1}$ and the masked ones of $y_{t}$. $\alpha$ is a margin parameter and $||\cdot||_2$ denotes the $\ell_2$ norm.

\noindent \textbf{Total Loss.}
In summary, we finalize the total loss as
\begin{small}
\begin{equation}
\label{loss3}
\begin{split}
    \mathcal{L} = \mathcal{L}_{dis} + \lambda_{tri}\mathcal{L}_{tri},
  \end{split}
\end{equation}
\end{small}
where $\lambda_{tri}$ denotes the balance paramneter, which is empirically set as 1 in our experiments. With Eq.(\ref{loss3}) to be minimized, $D$ can be trained to evaluate the semantic correlation, where the small score value from $D$ indicates that the correlation between the texture and structure is not strong, weakening the guidance of the structure.
To enhance the semantic correlation, an adaptive resampling strategy comes up, which is elaborated in the next section.

\subsubsection{Adaptive Resampling Strategy}
Different from the previous work \cite{lugmayr2022repaint} that directly performs the resampling strategy without the clear measurement, we devise an \emph{adaptive} resampling strategy, which regulates the semantic correlation between the texture (\emph{e.g.}, $y_{t-1}$) and structure (\emph{e.g.}, $x_{t-1}$) according to the score value from a discriminator in Sec.\ref{discriminator}. Specifically, when the score value of the semantic correlation between $y_{t-1}$ and $x_{t-1}$ is smaller than a specific threshold $\Delta$ (\emph{see the supplementary material for more intuitions}), we first obtain $\tilde{x}_{t}$ by adding the noise to $x_{t-1}$, then receive a updated $\tilde{x}_{t-1}$ by denoising $\tilde{x}_{t}$, which are described as follows:
\begin{equation}
\label{eq:resample}
\begin{split}
    &\tilde{x}_{t} = x_{t-1}+\delta_{t-1}(\mu_{x}-x_{t-1})dt+\sigma_{t-1}\mathrm{d}w,\\
    &\tilde{x}_{t-1} = \tilde{x}_{t}-(\mathrm{d} x_{t})_{\tilde{\epsilon}_{\varphi}}.
  \end{split}
\end{equation}
Based on Eq.(\ref{eq:resample}), $\tilde{x}_{t-1}$ is utilized to generate an updated $\tilde{y}_{t-1}$ via the trained noise network ${\tilde{\epsilon}}_{\phi}(y_{t},\tilde{x}_{t-1},t)$. Following that, we evaluate the correlation between $\tilde{y}_{t-1}$ and $\tilde{x}_{t-1}$ again, and repeat the above procedure; see Fig.\ref{adaptive_resampling}.

\section{Experiments}
\subsection{Implementation Details}
We experimentally evaluate StrDiffusion over three typical datasets, including: \textbf{Paris StreetView (PSV)} \cite{doersch2012makes} is a collection of street view images in Paris, consisting of 14.9k images for training and 100 images for validation; \textbf{CelebA} \cite{liu2015deep} is the human face dataset with 30k aligned face images, which is split into 26k training images and 4k validation images; \textbf{Places2}  \cite{zhou2017places} contains more than 1.8M natural images from varied scenes. The network is trained using $256 \times 256$ images with irregular masks \cite{liu2018image}. The noise-prediction network is constructed by removing group normalization layers and self-attention layers in the U-Net in  DDPM \cite{ho2020denoising} for inference efficiency. We adopt the Adam optimizer with $\beta_1$ = 0.9 and $\beta_2$ = 0.99. We set the timestep $T=100$ for the diffusion model. All the experiments are implemented with the pytorch framework and run on 4 NVIDIA 2080TI GPU.


\subsection{Why can StrDiffusion Work?}
In this section, the core goal is to confirm that the intuition of StrDiffusion --- the progressively sparse structure over time is beneficial to alleviate the semantic discrepancy issue between the masked and unmasked regions during the denoising process. We conduct the experiments on PSV and Places2 dataset. Fig.\ref{inpainting_process} illustrates that, as the denoising process progresses, the denoised results from IR-SDE \cite{luo2023image} always address the clear semantic discrepancy between the masked and unmasked regions (see Fig.\ref{inpainting_process}(a)); while for StrDiffusion, such discrepancy \emph{progressively} degraded until \emph{vanished}, yielding the consistent semantics (see Fig.\ref{inpainting_process}(b)), confirming our \emph{proposals} in Sec.\ref{optimal_solution} --- the guidance of the structure benefits the \emph{consistent semantics} between the masked and unmasked regions while the texture retains the \emph{reasonable semantics} in the inpainted results; \emph{see the supplementary material for more results.}

\subsection{Comparison with State-of-the-arts}
To validate the superiority of StrDiffusion, we compare it with the typical inpainting models, including: RFR \cite{li2020recurrent} adopts the CNNs to encoding the local information; PENNet\cite{yan2019PENnet}, HiFill\cite{yi2020contextual}, FAR\cite{cao2022learning}, HAN\cite{deng2022hourglass} and CMT\cite{Ko_2023_ICCV} globally associate the masked regions with unmasked ones via self-attention mechanism; MEDFE\cite{Liu2019MEDFE}, CTSDG\cite{Guo_2021_ICCV}, ZITS\cite{dong2022incremental} and DGTS\cite{liu2022delving} focus on the guidance of the structure,  while ignore the semantic consistency between the structure and texture; RePaint\cite{lugmayr2022repaint} and IR-SDE\cite{luo2023image} benefit from DDPMs whereas overlook the semantic consistence. For fair comparison, we reproduce RePaint by utilizing the pre-trained models from IR-SDE, denoted as RePaint*. Notably, the inpainted results are obtained by merging the masked regions from the denoised results with the original masked image.

\begin{figure}[t]
  \centering
    \setlength{\abovecaptionskip}{0.2cm}
\setlength{\belowcaptionskip}{-0.5cm}
  \includegraphics[width=0.9\linewidth]{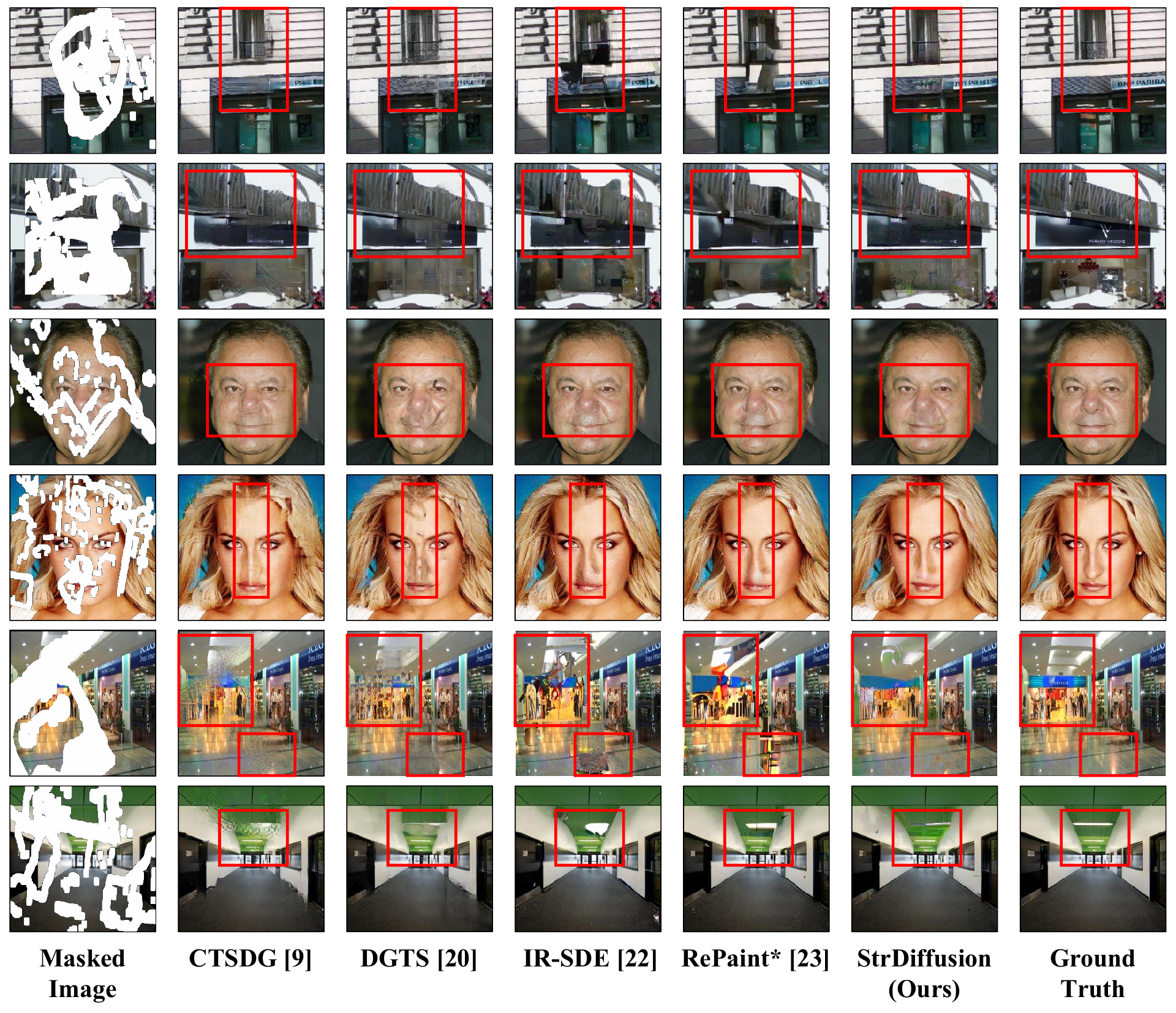}
  \caption{Comparison of the inpainted results with the state-of-the-arts under varied irregular masks on PSV, CelebA and Places2. Our StrDiffusion delivers the desirable inpainted results (marked as the red box) against others.}
  \label{visual_analysis}
\end{figure}

\begin{figure}[t]
    \setlength{\abovecaptionskip}{0.2cm}
\setlength{\belowcaptionskip}{-0.5cm}
  \centering
  \includegraphics[width=0.9\linewidth]{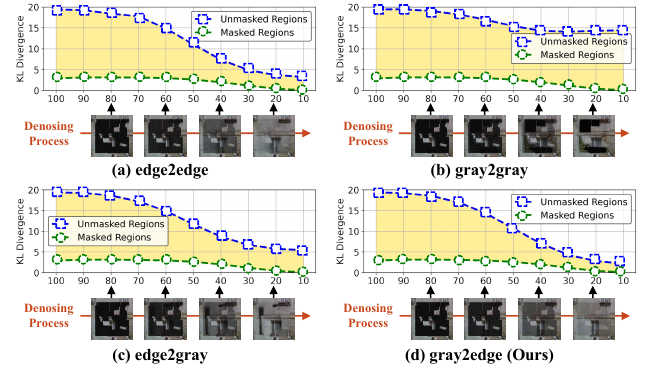}
  \caption{Ablation study about the the progressive sparsity for the structure over time on the PSV dataset. The KL divergence (lower is better) measures the distribution difference between the masked (unmasked) regions and the completed image (\emph{i.e.}, ground truth).}
  \label{varied_sparsity}
\end{figure}

\noindent \textbf{Quantitative and Qualitative analysis.}
The quantitative results in Table \ref{table:compare1} apart from the qualitative results in Fig.\ref{visual_analysis} summarize our findings below: StrDiffusion enjoys a much smaller FID score, together with larger PSNR and SSIM than the competitors, confirming that StrDiffusion effectively addresses semantic discrepancy between the masked and unmasked regions, while yielding the reasonable semantics.  Notably, RePaint* and IR-SDE still remain the large performance margins (at most 1.2\% for PSNR, 1.4\% for SSIM and 10\% for FID) compared to StrDiffusion, owing to the semantic discrepancy (e.g., the white wall in the row 1 of Fig.\ref{visual_analysis}) in the denoised results incurred by the dense texture.  Albeit MEDFE, CTSDG and ZITS focus on the guidance of the structure similar to StrDiffusion, they suffer from a performance loss due to the semantic discrepancy between the structure and texture, \emph{e.g.}, the woman's face for CTSDG in the row 4 of Fig.\ref{visual_analysis}, which further verifies our \emph{proposal} in Sec.\ref{tdim} --- \emph{the progressively sparse structure provides the time-dependent  guidance for texture denoising process}; \emph{see the supplementary material for more results.}

\subsection{Ablation Studies}
\subsubsection{Why Should the Semantic Sparsity of the Structure be Strengthened over Time?}
Recalling Sec.\ref{tdim}, the progressive sparsity for the structure is utilized to mitigate the semantic discrepancy, by setting the initial structure state as the \emph{grayscale} and the terminal structure state as the combination of the \emph{edge} and the noise during the structure diffusion process, denoted as \textbf{gray2edge}. To shed more light on the semantic sparsity of the structure, we replace the initial and terminal structure state, yielding several alternatives as \textbf{gray2gray}, \textbf{edge2edge} and  \textbf{edge2gray}; and perform the experiments on the PSV dataset. The structure sparsity of \textbf{gray2gray} and \textbf{edge2edge} is fixed over time.  Fig.\ref{varied_sparsity} illustrates that \textbf{gray2edge} (Fig.\ref{varied_sparsity}(d)) exhibits better consistency with meaningful semantics in the inpainted results against others, especially for \textbf{edge2gray}, implying the \emph{benefits} of strengthening structure sparsity over time. Notably, for \textbf{gray2gray}, the discrepancy issue in the denoised results still suffers (Fig.\ref{varied_sparsity}(b)), while \textbf{edge2edge} receives the poor semantics (Fig.\ref{varied_sparsity}(a)), which attributes to their \emph{invariant} semantic sparsity over time,  indicating \emph{time-dependent} impacts of the unmasked semantics on the texture denoising process (Sec.\ref{structure_matter}); \emph{see the supplementary material for more results.}
\begin{table}[t]
    \centering
    \caption{Ablation study about the adaptive resampling strategy on the PSV dataset. The best results are reported with \textbf{boldface}. }
    \resizebox{0.9\linewidth}{1.1cm}{
    \begin{tabular}{@{}c||ccc|ccc@{}}
    \toprule
    \multirow{2}*{\textbf{Cases} }
    &\multicolumn{3}{c}{\textbf{Hard mask}} &\multicolumn{3}{c}{\textbf{Easy mask}}\\ \cmidrule(r){2-4} \cmidrule(r){5-7}
    &\textbf{PSNR}$\uparrow$ &\textbf{SSIM}$\uparrow$ &\textbf{FID}$\downarrow$ &\textbf{PSNR}$\uparrow$ &\textbf{SSIM}$\uparrow$ &\textbf{FID}$\downarrow$\\
    \hline\hline
    \textbf{A} &24.65 &0.851 &55.73 &27.69 &0.913 &33.73\\
    \hline
    \textbf{B} &25.47 &0.863 &46.44 &28.31 &0.921 &28.43 \\
    \textbf{C} &25.59  &0.867 &47.31 &28.64 &0.925 &28.20\\
    \hline
    \textbf{D} &\textbf{25.94}  &\textbf{0.875} &\textbf{42.28} &\textbf{28.85} &\textbf{0.929} &\textbf{26.74} \\
\bottomrule
    \end{tabular}
  \label{table:cross}
  }
  \vspace{-0.5cm}
\end{table}

\subsubsection{How does Adaptive Resampling Strategy Benefit Inpainting?}
To verify the effectiveness of the adaptive resampling strategy (Sec.\ref{further_insight}), we perform the ablation study on the PSV dataset from the following cases: \textbf{A}:  refining the inpainted results of IR-SDE\cite{luo2023image} via the resampling strategy in Repaint; \textbf{B}: removes the discriminator, which performs the adaptive resampling strategy without considering the semantic correlation between the texture and structure; \textbf{C}:  StrDiffusion without the adaptive resampling strategy; \textbf{D}: the proposed StrDiffusion. Table \ref{table:cross} suggests that case \textbf{D} delivers great gains (at most 13.45\%) to other cases, which is \emph{evidenced} that the adaptive resampling strategy succeeds in regulating the semantic correlation for the desirable results (Sec.\ref{further_insight}); case \textbf{D} upgrades beyond case \textbf{A} despite of the combination with the resampling strategy, which, in turn, verifies the \emph{necessity} of the time-dependent guidance from the structure.  In particular, case \textbf{B} bears a large performance degradation (at most 4.16\%), confirming the importance of \emph{adaptively} resampling strategy as per the semantic correlation between the structure and texture in Sec.\ref{discriminator}.

%

\section{Conclusion}
In this paper, we propose a structure-guided diffusion model for inpainting (StrDiffusion), to reformulate the conventional texture denoising process under the guidance of the structure to derive a simplified denoising objective; while revealing: 1) the semantics from the unmasked regions essentially offer the time-dependent guidance for the texture denoising process; 2) the progressively sparse structure well tackle the semantic discrepancy in the inpainted results. Extensive experiments verify the superiority of StrDiffusion to the state-of-the-arts.\\
\textbf{Acknowledgments} This work was supported by the National Natural Science Foundation of China (U21A20470, 62172136, 72188101), the Institute of Advanced Medicine and Frontier Technology under (2023IHM01080).

{
    \small
    \bibliographystyle{ieeenat_fullname}
    \bibliography{main}
}

\appendix
\clearpage
\setcounter{page}{1}
\maketitlesupplementary

Due to page limitation of the main body, as indicated, the supplementary material offers more details on the ideal reverse state $\tilde{y}_{t-1}^{*}$, further discussion on the threshold $\Delta$, additional quantitative results and more visual results with higher resolution, which are summarized below:
\begin{itemize}
    \item More derivation details on the ideal reverse state $\tilde{y}_{t-1}^{*}$, as mentioned in Sec.2.2.2 of the main body (Sec.\ref{sec0}).
    \item More \emph{intuition} on the threshold $\Delta$ involved in the adaptive resampling strategy, as mentioned in Sec.2.4.2 of the main body (Sec.\ref{sec1}).
    \item Visualization of the denoised results with \emph{higher resolution} for IR-SDE \cite{luo2023image} and StrDiffusion during the denoising process, as mentioned in Sec.3.2 of the main body (Sec.\ref{sec2}).
    \item Additional quantitative results for the comparison with state-of-the-arts, as mentioned in Sec.3.3 of the main body (Sec.\ref{sec3}).
    \item Additional visual results about the ablation study about the progressive sparsity for the structure over time, as mentioned in Sec.3.4.1 of the main body (Sec.\ref{sec4}).
\end{itemize}

\section{More details on the Ideal Reverse State $\widetilde{y}_{t-1}^{*}$}
\label{sec0}
 Due to page limitation, we offer more derivation details from Eq.(7) to Eq.(11) in the main body. Based on the Eq.(7) of the main body, the optimal reverse state is naturally acquired by minimizing the negative log-likelihood:
\begin{equation}
\label{}
\begin{split}
    \tilde{y}_{t-1}^{*}&=\arg\min_{y_{t-1}} [-\log q(y_{t-1}|y_{t},y_{0},x_{t-1},x_{0})]\\[6pt]
    &=\arg\min_{y_{t-1}} [-\log \frac{q(y_{t-1}|y_{0})}{q(x_{t-1}|x_{0})}],
    \end{split}
\end{equation}
where $\tilde{y}_{t-1}^{*}$ denotes the ideal state reversed from $\tilde{y}_{t}$ under the structure guidance.
To solve the above objective, we compute its gradient as:
\begin{equation}
\label{eq:ideal1}
\begin{split}
    &\quad\nabla_{\tilde{y}_{t-1}^{*}} \left\{-\log q(\tilde{y}_{t-1}^{*}|y_{t},y_{0},x^{*}_{t-1},x_{0})\right\} \\[6pt]
    &=\nabla_{\tilde{y}_{t-1}^{*}} \left\{-\log \frac{q(\tilde{y}_{t-1}^{*}|y_{0})}{q(x_{t-1}^{*}|x_{0})}\right\} \\[6pt]
     &=-\nabla_{\tilde{y}_{t-1}^{*}}\log q(\tilde{y}_{t-1}^{*}|y_{0})+\nabla_{x_{t-1}^{*}}\log q(x_{t-1}^{*}|x_{0})\\[6pt]
     &=\frac{\tilde{y}_{t-1}^{*}-\mu_{y}-e^{-\overline{\theta}_{t-1}}({y}_{0}-\mu_{y})}{1-e^{-2\overline{\theta}_{t-1}}} \\[6pt]
     &\quad- \frac{{x}_{t-1}^{*}-\mu_{x}-e^{-\overline{\delta}_{t-1}}({x}_{0}-\mu_{x})}{1-e^{-2\overline{\delta}_{t-1}}},
    \end{split}
\end{equation}
where the texture $\mu_{y}$ is the masked version of its initial state $y_{0}$ and $\theta_{t}$ is time-dependent parameter that characterizes the speed of the mean-reversion, the structure $\mu_{x}$ is the masked version of its initial state $x_{0}$ and $\delta_{t}$ is time-dependent parameter that characterizes the speed of the mean-reversion, $\overline{\theta}_{t-1}=\int_{0}^{t-1}\theta_{z}\mathrm{d}z$ and $\overline{\delta}_{t-1}=\int_{0}^{t-1}\delta_{z}\mathrm{d}z$. Setting Eq.(\ref{eq:ideal1}) to be zero, we can get $\tilde{y}_{t-1}^{*}$ as:
\begin{equation}
\begin{split}
   \tilde{y}_{t-1}^{*}= &\frac{(1-e^{-2\overline{\theta}_{t-1}})({x}_{t-1}^{*}-\mu_{x})}{1-e^{-2\overline{\delta}_{t-1}}} \\[6pt] &-\frac{(1-e^{-2\overline{\theta}_{t-1}})e^{-\overline{\delta}_{t-1}}({x}_{0}-\mu_{x})}{1-e^{-2\overline{\delta}_{t-1}}} \\[6pt]
   &+e^{-\overline{\theta}_{t-1}}({y}_{0}-\mu_{y}) + \mu_{y},
    \end{split}
\end{equation}
where $x_{t-1}^{*}$ is the ideal state reversed from $x_{t}$ for the structure, given as:
\begin{equation}
\begin{split}
    x_{t-1}^{*}=& \frac{1- e^{-2\overline{\delta}_{t-1}}}{1- e^{-2\overline{\delta}_{t}}}e^{-\delta_{t}^{'}}{(x_{t}-\mu_{x})}\\[6pt]
    &+\frac{1- e^{-2\delta_{t}^{'}}}{1- e^{-2\overline{\delta}_{t}}}e^{-\overline{\delta}_{t-1}}{(x_{0}-\mu_{x})}+ {\mu_{x}}.
    \end{split}
\end{equation}
To simplify the notation, $\delta_{t}^{'}=\int_{t-1}^{t}\delta_{i}\mathrm{d}i$, we can derive  the ideal reverse state $\tilde{y}_{t-1}^{*}$ as:
\begin{equation}
\begin{split}
    \tilde{y}_{t-1}^{*}=& \frac{1- e^{-2\overline{\theta}_{t-1}}}{1- e^{-2\overline{\delta}_{t}}}e^{-\delta_{t}^{'}}(x_{t}-\mu_{x}) \\
    &+\frac{(1- e^{-2\overline{\theta}_{t-1}})(e^{-2\overline{\delta}_{t}}-e^{-2\delta_{t}^{'}})}{(1- e^{-2\overline{\delta}_{t-1}})(1- e^{-2\overline{\delta}_{t}})}e^{-\overline{\delta}_{t-1}}  (x_{0}-\mu_{x})\\
    &+e^{-\overline{\theta}_{t-1}}(y_{0}-\mu_{y})+ \mu_{y}.
    \end{split}
\label{y_s}
\end{equation}
Since $\overline{\delta}_{t} = \overline{\delta}_{t-1} + \delta_{t}^{'}$, we can reformulated the second term in Eq.(\ref{y_s}) as follows:
\begin{equation}
\begin{split}
    &\frac{(1- e^{-2\overline{\theta}_{t-1}})(e^{-2\overline{\delta}_{t}}-e^{-2\delta_{t}^{'}})}{(1- e^{-2\overline{\delta}_{t-1}})(1- e^{-2\overline{\delta}_{t}})}e^{-\overline{\delta}_{t-1}}  (x_{0}-\mu_{x})\\
    =&\frac{(1- e^{-2\overline{\theta}_{t-1}})(e^{-2(\overline{\delta}_{t-1} + \delta_{t}^{'})}-e^{-2\delta_{t}^{'}})}{(1- e^{-2\overline{\delta}_{t-1}})(1- e^{-2\overline{\delta}_{t}})}e^{\delta_{t}^{'}-\overline{\delta}_{t}}  (x_{0}-\mu_{x})\\
    =&\frac{(1- e^{-2\overline{\theta}_{t-1}})(e^{-2\overline{\delta}_{t-1}}-1)}{(1- e^{-2\overline{\delta}_{t-1}})(1- e^{-2\overline{\delta}_{t}})}e^{-2\delta_{t}^{'}}e^{\delta_{t}^{'}-\overline{\delta}_{t}}  (x_{0}-\mu_{x})\\
    =&-\frac{(1- e^{-2\overline{\theta}_{t-1}})}{(1- e^{-2\overline{\delta}_{t}})}e^{-\delta_{t}^{'}}e^{-\overline{\delta}_{t}}  (x_{0}-\mu_{x})\\
    =&-\left(\frac{(1- e{}^{-2\overline{\theta}_{t-1}})}{(1- e^{-2\overline{\delta}_{t}})}e^{-\delta_{t}^{'}}\right)e^{-\overline{\delta}_{t}}  (x_{0}-\mu_{x}).
    \end{split}
\end{equation}
Based on the above, we have
\begin{equation}
\begin{split}
    \tilde{y}_{t-1}^{*}=&\underbrace{  \left(\frac{1- e^{-2\overline{\theta}_{t-1}}}{1- e^{-2\overline{\delta}_{t}}}e^{-\delta_{t}^{'}}\right)   (x_{t}-\mu_{x})}_{\text{Consistency term for masked regions}}\\
    &-\underbrace{ \left(\frac{1- e^{-2\overline{\theta}_{t-1}}}{1- e^{-2\overline{\delta}_{t}}}e^{-\delta_{t}^{'}}\right)e^{-\overline{\delta}_{t}}  (x_{0}-\mu_{x})}_{\text{Balance term for masked regions}}\\
    & +\underbrace{e^{-\overline{\theta}_{t-1}} (y_{0}-\mu_{y})}_{\text{Semantics  term for masked regions}} + \underbrace{\mu_{y}}_{\text{Unmasked regions}}.
    \end{split}
\end{equation}

\begin{algorithm}[]  
	\caption{Adaptive Resampling Strategy}
    \label{algorithm}
	\LinesNumbered 
    \KwIn{the noise version of masked texture $y_{T}$ and the noise version of masked structure $x_{T}$, trained noise-prediction networks ${\tilde{\epsilon}}_{\phi}$ and ${\tilde{\epsilon}}_{\varphi}$ for the texture and structure,  the timestep $T$, the discriminator $D$, the maximum number of iterations $U$}
	\KwOut{The denoised inpainted result $y_{0}$}

    \For{t = T, \ldots , 1}{
		Denoised structure ${x}_{t-1} = {x}_{t}-(\mathrm{d}{x}_{t})_{{\tilde{\epsilon}}_{\varphi}(x_{t},t)}$

		Denoised texture ${y}_{t-1} = {y}_{t}-(\mathrm{d}{y}_{t})_{{\tilde{\epsilon}}_{\phi}({y}_{t},x_{t-1},t)}$

        Obtain the threshold $\Delta = {D} ({y}_{t-1},x_{t-1},t-1)$

        \For{u = 1, \ldots , U}{
                 $\tilde{x}_{t} = x_{t-1}+(\mathrm{d}{x}_{t-1})$

                 $\tilde{x}_{t-1} = \tilde{x}_{t}-(\mathrm{d}{\tilde{x}}_{t})_{{\tilde{\epsilon}}_{\varphi}(\tilde{x}_{t},t)}$

            		$\tilde{y}_{t-1} = {y}_{t}-(\mathrm{d}{y}_{t})_{{\tilde{\epsilon}}_{\phi}(\tilde{y}_{t},\tilde{x}_{t-1},t)}$

                    Obtain the score $S = {D} (\tilde{y}_{t-1},\tilde{x}_{t-1},t-1)$

                \eIf{$S < \Delta$}{

                        Update ${x}_{t-1} = \tilde{x}_{t-1}$

                        Update ${y}_{t-1} = \tilde{y}_{t-1}$

    		}{
    			break
    		}
        }

	}
 \Return the denoised results $y_0$
\end{algorithm}

\section{More Intuition on the Threshold $\Delta$ in the Adaptive Resampling Strategy}
\label{sec1}
The specific threshold $\Delta$ in the adaptive resampling strategy is utilized to evaluate the semantic correlation between the structure and texture during the inference denoising process. Specifically, when the score value $S$ from the discriminator $D$ is smaller than the  threshold $\Delta$, \emph{i.e.},  $S<\Delta$, we will perform the resampling operation for the structure to enhance the semantic correlation for desirable results. A naive strategy is to mutually select a fixed value of $\Delta$, which, however, is inflexible, since the semantic correlation actually \emph{varies} greatly as the inference denoising process progresses. Unlike the previous work \cite{lugmayr2022repaint} that aims to condition the denoising process for image inpainting task via the resampling strategy, our adaptive resampling strategy actually serves as a by-product; the goal  is to refine the correlation between the structure and texture as possible.
To this end, we present to exploit the structure $x_{t-1}$ and the texture $y_{t-1}$ without the adaptive resampling strategy in the $t$-th timestep, to serve as the inputs for the discriminator $D$, leading to a score value, which is treated as the threshold $\Delta$. Under such case, the semantic correlation between the structure and texture can be always boosted to yield better denoised results.
Based on the threshold $\Delta$, the whole adaptive resampling strategy is summarized  in Algorithm \ref{algorithm}.
\begin{table*}[t]
    \centering
    \caption{Comparison of quantitative results (\emph{i.e.}, PSNR, SSIM, and FID) under varied mask ratios on CelebA with irregular mask dataset. $\uparrow$: Higher is better; $\downarrow$: Lower is better. The best results are reported with \textbf{boldface}.}
    \label{table:compare2}
    \resizebox{0.8\linewidth}{!}{
    \begin{tabular}{@{}lc|ccc|ccc|ccc@{}}
    \toprule
    \multicolumn{2}{c}{\textbf{Metrics}} & \multicolumn{3}{c}{\textbf{PSNR$\uparrow$}} & \multicolumn{3}{c}{\textbf{SSIM$\uparrow$}} & \multicolumn{3}{c}{\textbf{FID$\downarrow$}} \\
    \cmidrule(lr){1-2} \cmidrule(lr){3-5} \cmidrule(lr){6-8} \cmidrule(lr){9-11}
    \textbf{Method} & \textbf{Venue} & \textbf{0-20\%} & \textbf{20-40\%} & \textbf{40-60\%} & \textbf{0-20\%} & \textbf{20-40\%} & \textbf{40-60\%} & \textbf{0-20\%} & \textbf{20-40\%} & \textbf{40-60\%} \\
    \midrule
    \textit{PIC} \cite{zheng2019pluralistic} & CVPR' 19 & 33.67 & 26.48 & 21.58 & 0.978 & 0.934 & 0.865 & 2.340 & 6.430 & 14.22  \\
    \hline
    \textit{MAT} \cite{li2022mat} & CVPR' 22 & 35.31 & 27.76 & 23.22  & 0.984 & 0.946 & 0.888 & 0.900 & 2.550 & \textbf{4.600} \\
    \textit{CMT} \cite{Ko_2023_ICCV} & ICCV' 23 & 35.92 & 28.24 & 23.78  & 0.986 & 0.952 & 0.900  & 0.840 & 2.540 & 5.230 \\
    \hline
    \textit{ICT} \cite{wan2021high} & CVPR' 21 & 33.27 & 26.40 & 21.84 & 0.979 & 0.939 & 0.877 & 1.870 & 5.610 & 12.42 \\
    \textit{BAT} \cite{yu2021diverse} & MM' 21 & 34.63 & 26.91 & 22.26  & 0.983 & 0.944 & 0.883  & 1.060 & 3.750 & 7.300  \\
    \hline
    \textit{RePaint*} \cite{lugmayr2022repaint} & CVPR' 22 & 36.23 & 29.01 & 23.92  & 0.991 & 0.969 & 0.912  & 0.790 & 2.530 & 5.030  \\
    \textit{IR-SDE} \cite{luo2023image} & ICML' 23 & 36.01 & 28.85 & 23.76 & 0.991 & 0.966 & 0.910  & 0.870 & 2.840 & 5.700 \\
    \hline
    \hline
    \textbf{StrDiffusion (Ours)} & - & \textbf{36.44} & \textbf{29.31} & \textbf{24.50}  & \textbf{0.994} & \textbf{0.971} & \textbf{0.923}  & \textbf{0.660} & \textbf{2.400} & 4.950 \\
    \bottomrule
    \end{tabular}
    }
\end{table*}

\section{Visualization of the Denoised Results with Higher Resolution }
\label{sec2}
\begin{figure*}
  \centering
  \includegraphics[width=0.9\linewidth]{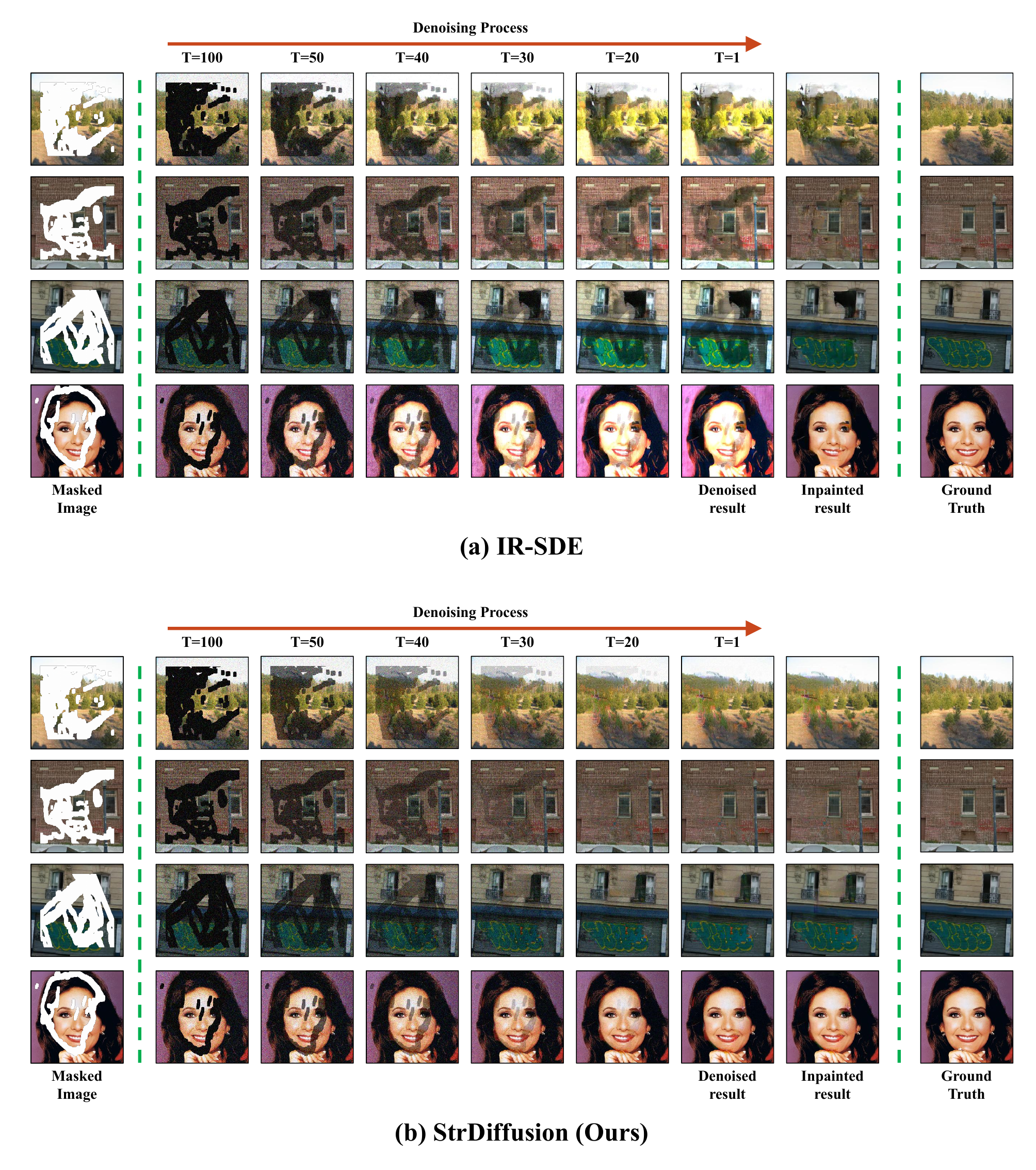}
  \caption{Visualization of the denoised results for IR-SDE (a) and StrDiffusion (b) in the varied timesteps during the denoising process, as an \emph{extension} of Fig.6 in the main body.}
  \label{consistency-supp}
\end{figure*}

As mentioned in Sec.3.2 of the main body, due to page limitation, we further provide more denoised results with \emph{higher resolution} on the Places2,  PSV and CelebA datasets; see Fig.\ref{consistency-supp}. The results show that, unlike the denoised results from IR-SDE \cite{luo2023image} that always address the \emph{clear} semantic discrepancy between the masked and unmasked regions (see Fig.\ref{consistency-supp}(a)), for StrDiffusion, such discrepancy progressively degraded until \emph{vanished}, yielding the consistent semantics (see Fig.\ref{consistency-supp}(b)), which are consistent with our analysis in the main body.


\section{Additional Quantitative Results}
\label{sec3}
\textbf{Evaluation metric.} We adopt three metrics to evaluate the inpainted results below: 1) peak signal-to-noise ratio (PSNR); 2) structural similarity index (SSIM) \cite{wang2004image}; and 3) Fréchet Inception Score (FID) \cite{heusel2017gans}. PSNR and SSIM are used to compare the low-level differences over pixel level between the generated image and ground truth. FID evaluates the perceptual quality by measuring the feature distribution distance between the synthesized and real images.

As indiacted in the main body, we further exhibit additional quantitative results under varied mask ratios on CelebA with irregular mask dataset; see Table.\ref{table:compare2}.
It is observed that our StrDiffusion enjoys a much smaller FID score, together with larger PSNR and SSIM than the competitors, confirming that StrDiffusion effectively addresses semantic discrepancy between the masked and unmasked regions, while yielding the reasonable semantics. Notably, RePaint* and IR-SDE still remain the large performance margins (at most 1.0\% for PSNR, 0.2\% for SSIM and 5.2\% for FID) compared to StrDiffusion, owing to the semantic discrepancy in the denoised results incurred by the dense texture. Albeit ICT and BAT focus on the guidance of the structure similar to StrDiffusion, they suffer from a performance loss due to the semantic discrepancy between the structure and texture, which \emph{confirms} our proposal in Sec.2.2 of the main body --- \emph{the progressively sparse structure provides the time-dependent  guidance for texture denoising process}.

\section{Additional Visual Results for the Ablation Study in Sec.3.4.1 of the Main Body}
\label{sec4}
\begin{figure*}
  \centering
  \includegraphics[width=\linewidth]{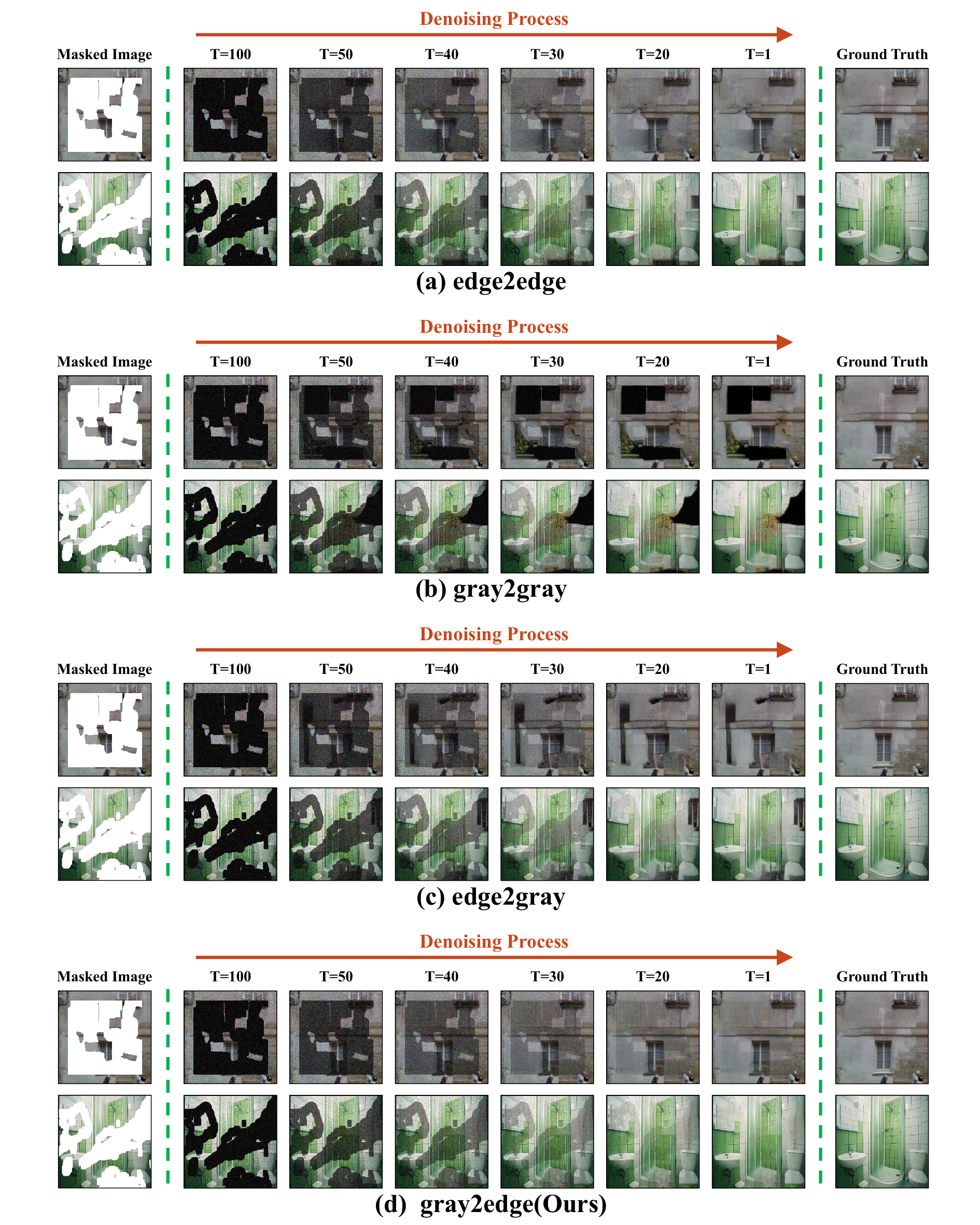}
  \caption{Additional visual results for the ablation study about the progressive sparsity for the structure over time, as an \emph{extension} of Fig.8 in the main body.}
  \label{which-supp}
\end{figure*}

The ablation study in Sec.3.4.1 of the main body aims to verify why the semantic sparsity of the structure should be strengthened over time. In this section, we further exhibit additional visual results by performing the experiments on the PSV and Places2 datasets; see Fig.\ref{which-supp}. It is observed that our \textbf{gray2edge} (Fig.\ref{which-supp}(d)) exhibits better consistency with meaningful semantics in the inpainted results against others, especially for \textbf{edge2gray} (Fig.\ref{which-supp}(c)), implying the \emph{benefits} of strengthening the sparsity of the structure over time. Notably, for \textbf{gray2gray}, the discrepancy issue in the denoised results still suffers (Fig.\ref{which-supp}(b)), while \textbf{edge2edge} receives the poor semantics (Fig.\ref{which-supp}(a)), which attributes to their \emph{invariant} semantic sparsity over time.
Such fact \emph{confirms} our proposals in Sec.3.4.1 of the main body.

\end{document}